# Unlocking Your Sales Insights: Advanced XGBoost Forecasting Models for Amazon Products

Meng Wang[1], Yuchen Liu[1] [0009-0007-1525-7676], Gangmin Li[2][0000-0003-4006-7472], Terry R. Payne[3][0000-0002-0106-8731], Yong Yue[1][0000-0001-7695-4538], Ka Lok Man[1*] [0000-0002-5787-4716]

[1] Dept. of Computing, Xi'an Jiaotong-Liverpool University
[2] College of Industry-Entrepreneurs (CIE), Xi'an Jiaotong-Liverpool University
[3] Dept. of Computer Science, University of Liverpool
Meng.Wang22@student.xjtlu.edu.cn,
Yuchen.Liu21@student.xjtlu.edu.cn,
gangmin.li@xjtlu.edu.cn, T.R.Payne@liverpool.ac.uk,
Yong.Yue@xjtlu.edu.cn, Ka.Man@xjtlu.edu.cn

**Abstract.** One of the important factors of profitability is the volume of transactions. An accurate prediction of the future transaction volume becomes a pivotal factor in shaping corporate operations and decision-making processes. E-commerce has presented manufacturers with convenient sales channels to, with which the sales can increase dramatically. In this study, we introduce a solution that leverages the XGBoost model to tackle the challenge of predicting sales for consumer electronics products on the Amazon platform. Initially, our attempts to solely predict sales volume yielded unsatisfactory results. However, by replacing the sales volume data with sales range values, we achieved satisfactory accuracy with our model. Furthermore, our results indicate that XGBoost exhibits superior predictive performance compared to traditional models.

**Keywords:** Ensemble Learning, Consumer Electronics, Sales Forecasting, Amazon, XGBoost, CatBoost, GBDT, E-commerce First Section

## 1 Introduction

With the growth of the Internet and e-commerce platforms, e-commerce has become a key sales channel for manufacturers. Accurate sales forecasts on e-commerce platforms can help dealers and manufacturers cut inventory costs, boost profits, and adjust pricing strategies in time [1, 2]. Such forecasts also assist upstream suppliers in anticipating the demand for finished goods, lessening the demand distortion on the supply chain [3]. Accurate sales forecasting is vital for marketing strategies, efficient business operations, and profitability.

Many studies have achieved excellent results for commodity sales forecasting using different technologies and models. Numerous scholars have made substantial progress in forecasting accuracy by refining existing time series models or adopting a multi-model integration approach [4, 5]. Conversely, the emergence of machine learning and deep learning techniques has showcased remarkable proficiency in addressing scenarios with limited historical data or subtle patterns [6, 7]. A rigorous study conducted by Dipillo and colleagues compared the predictive power of the support vector machine (SVM) against the ARIMA model using a standardised dataset for forecasting transaction values [8]. The results demonstrated SVM's potential in navigating the intricacies of sales forecasting, especially in scenarios where historical data is scarce or exhibits unpredictable fluctuations. Additionally, the numerous benefits of machine learning and deep learning techniques have sparked a significant amount of research examining the impact of product features on sales and developing methods for forecasting e-commerce revenue [9, 10]. The factors that impact sales, such as holiday periods [11, 12], contribute significantly to e-commerce sales forecasting. Hence, including multidimensional and multifaceted features is crucial in generating accurate predictions.

The sales data employed in this study is sourced from the Amazon website, particularly consumer electronics products. We aim to predict sales volumes by leveraging XGBoost algorithms with this dataset. Furthermore, these sales volumes are communicated as range estimates instead of specific values, a factor that potentially undermines the model's predictive precision. Considering these limitations, this study further subdivides the sales volume data based on the initial dataset to evaluate the model's practical reliability in short-term sales forecasting. The remainder of this paper is organised as follows: Section Two details the methodology, including data sources, preprocessing, and model implementation. Section Three presents the experimental results, focusing on XGBoost's predictions in consumer electronics and comparing them to other models. Lastly, Section Four concludes by discussing limitations and outlining future research work.

## 2    Methodology

XGBoost is an additive model aggregating multiple decision tree predictive outcomes. Specifically, it comprises an ensemble of k-base learners. The overall model can be expressed as follows:

$$\hat{y}_i = \sum_{k=1}^{K} f_k(x_i), f_k \in F \tag{1}$$

In formulation (1), the $K$ represents the number of the trees; $f_k$ is a specific tree; and $\hat{y}_i$ is the model output for a given sample length $n$ and the number of features $m$, then:

$$D = \{(x_i, y_i)\}(|D| = n, x_i \in R^m, y_i \in R) \tag{2}$$

In equation (2), $x_i$ represents the input of the $i$ sample; $y_i$ represents the output corresponding to the input of sample $i$; F is the space of the CART tree, which can be expressed as:

$$F = \{f(x) = w_q(x)\}(q: R^{\wedge}m \to T, w \in R^{\wedge}T) \tag{3}$$

3where, q represents the structure of the tree; T is the child node of the tree; $f(x)$ is the CART tree structure $q$ and child node weight $w$. By incorporating the regularisation term into the objective function:

$$Obj = \sum_i l(\hat{y}_i, y_i) + \sum_k \Omega(f_k) \tag{4}$$

$$\Omega(f) = rT + \frac{1}{2}\lambda ||w||^2 \tag{5}$$

In equation (4), $\hat{y}_i$ and $y_i$ represent the predicted value and actual label value of the $i$-th sample respectively; $\gamma$ and $\lambda$ are weight coefficients; $T$ is the total number of leaves; $w$ represents the model leaf node weight. $Obj$ is the objective function, $\sum_i l(\hat{y}_i, y_i)$ is the loss error which represents the error between the predicted value and the true value and the second half is the regular term.

## 3  Experiment

### 3.1  Data

The datasets employed in this study were derived from product information on the Amazon website, comprising 1565 distinct products. This comprehensive dataset encompasses various attributes such as product details, brand information, colour variations, manufacturer specifications, item weights, pricing details, customer ratings, review counts, sales volumes, and logistics timeframes. The primary product categories represented in the dataset include wireless headphones, gaming keyboards, computer mice, and air fryers. A detailed breakdown of the feature definitions is presented in Table 1.

**Table 1.** Feature description.

| Num | Feature Name | Feature Meaning | Type |
|---|---|---|---|
| 1 | Products | Commodity classification | object |
| 2 | Brand | Brand of Commodity & object | object |
| 3 | Colour | Color of Commodity & object | object |
| 4 | Manufacturer | Manufacturer of Commodity | object |
| 5 | Price | Price of Commodity | object |
| 6 | Rating | The customer's rating of the Commodity | float |
| 7 | Number of Rating | The number of customer reviews | float |
| 8 | Shipment | Shipment & Time spent on logistics (day) | float |
| 9 | Weight Pounds | Weight Pounds & The weight of Commodity (pounds) | float |

### 3.2  Feature Engineering

This study employs several strategies tailored to different features to handle missing values. Firstly, for features where a missing value genuinely represents zero, such as ratings, review counts, and sales, all missing entries are replaced with zero to ensure practical reliability. Secondly, when encountering missing brand or manufacturer data (which often overlap), information from the other attributes is utilised to fill the gap. For missing logistics time and product weight values, the initial approach involves substituting them with the average values of similar products from the same brand and type. If such data is unavailable, the average values of products within the same type are used. Regarding colour features, specific shade descriptors (e.g., "dark grey") are simplified to their generic counterparts (e.g., "grey"). Products exhibiting multiple colours are categorised accordingly: dual-coloured products are labelled composite, while those with three or more colours are designated as "colourful". Furthermore, this study applies one-hot encoding to categorical variables such as categories, brands, colours, and manufacturers. The dataset is partitioned for prediction purposes into a training set comprising 80% of the data and a test set containing the remaining 20%.

### 3.3 Sales prediction

The initial prediction evaluation in this article focuses on forecasting sales volume using the original dataset. While the outcomes were not optimal, with all five models exhibiting an MSE exceeding 200,000, it is noteworthy that XGBoost demonstrated superior performance compared to traditional machine learning models. The evaluation of the model's predictive capabilities posed a challenge due to the considerable magnitude of the MSE. Consequently, we undertook enhancements based on XGBoost to mitigate the excessive MSE stemming from the original data. However, it should be acknowledged that inaccuracies in the original dataset about product sales volume might have influenced these results. Hence, enhancing the quality of the original data is paramount for boosting model prediction proficiency.

Given the ambiguous data presented on the Amazon website, this study aims to alleviate bias in model predictions by categorising sales volume into specific ranges. Specifically, we have segmented the sales volume into eight groups: 0-50, 50-100, 100-300, 500-1000, 1000-3000, 3000-5000, and 5000-10000. Concurrently, we substituted the original dataset's sales volume data with these newly generated range values and reintroduced them into the model for predictive purposes. The predicted outcomes of the range values are documented in Table 2.

**Table 2.** Forecasting result for range.

| Model | MSE | RMSE | MAE |
| --- | --- | --- | --- |
| GBDT | 3.45 | 1.86 | 1.23 |
| XGBoost | 1.93 | 1.39 | 1.07 |
| Linear | 1.06E+14 | 1.03E+07 | 2.29E+06 |
| Bayes | 4.47 | 2.11 | 1.66 |
| SVM | 3.52 | 1.88 | 1.51 |

The results documented in Table 2 reveal substantial enhancements in the predictive performance of each model when applied to range values as opposed to ordinary values. Regarding MSE, GBDT attained a value of 3.45, Bayes achieves 4.67, and SVM yields 3.51. However, XGBoost emerges as the superior model, exhibiting an MSE of only 1.93. This outstanding performance indicates that the average deviation between the predicted and actual values for XGBoost is less than two ranges. Furthermore, XGBoost showcases its dominance in prediction accuracy by attaining the lowest RMSE (1.389) and MAE (1.072) among all five models considered. These findings underscore the exceptional predictive capabilities of XGBoost, characterised by minimal average prediction errors. The linear regression exhibited relatively unusual results in this predictive experiment, with all three performance indicators exceeded 2,000,000. This observation suggests that the processing proficiency of linear regression in non-time series problems is considerably inferior to our enhanced XGBoost model. The RMSE metric further supports XGBoost's superiority over the other four models in addressing significant error problems. Given the relatively small size of our dataset and its vulnerability to overfitting and extreme values, the performance of the alternative models in terms of RMSE needs to improve compared to XGBoost's proficiency in this area. Similarly, this holds for MAE as well. In conclusion, XGBoost performs remarkably in predicting electronic consumer product sales, highlighting its utility and effectiveness.

## 4 Conclusion

The current research utilises a dataset of 1,565 distinct product data entries spanning four unique categories from the Amazon website. This study incorporates various attributes of these products' non-time series to formulate a suitable dataset. Subsequently, this dataset is employed in an XGBoost model to forecast sales volume and range. A comparative analysis with traditional machine learning models assessed the model's predictive performance. The findings reveal that the XGBoost model demonstrates superior performance across three primary evaluation metrics: Mean Squared Error (MSE), Root Mean Squared Error (RMSE), and Mean Absolute Error (MAE). This underscores the model's reliability and effectiveness in predicting sales outcomes, particularly for consumer electronics products characterised by rapid product iteration and low short-term repurchase rates.

Despite achieving promising results with the methodology employed in this study, it is essential to acknowledge certain limitations. This study focused on specific commodity categories with potential substitutes, such as air fryers vs. conventional fryers. Future research should aim for a broader scope, particularly in electronics, to enhance the reliability of sales predictions. In future research directions, the primary objective will be to augment the dataset with additional product entries and expand the scope of product categories. Furthermore, integrating other machine learning models with XGBoost holds promise for developing a more reliable and robust predictive framework. Lastly, future studies aim to explore the correlation between features and prediction outcomes to construct a more efficient and accurate prediction model.

## Acknowledgement

This work is partially supported by the XJTLU AI University Research Centre and Jiangsu Province Engineering Research Centre of Data Science and Cognitive Computation at XJTLU. Also, it is partially funded by the Suzhou Municipal Key Laboratory for Intelligent Virtual Engineering (SZS2022004) as well as funding: XJTLU-REF-21-01-002 and XJTLU Key Program Special Fund (KSF-A-17).